\relax
\documentclass[letterpaper]{article} 

\usepackage{aaai20}  
\usepackage{times}  
\usepackage{helvet} 
\usepackage{courier}  
\usepackage[hyphens]{url}  
\usepackage{graphicx} 
\usepackage{graphicx}  
\usepackage{boxedminipage}
\usepackage{algorithm}
\usepackage{algorithmic}
\usepackage{amsmath}
\usepackage{graphicx}
\usepackage{subfigure}
\usepackage{multirow} 
\usepackage{makecell}
\usepackage{color}
\usepackage{epsfig}
\usepackage{amsmath}
\usepackage{amssymb}
\usepackage{rotating}
\def\mvp{\vspace*{-0.1in}}

\newcommand{\seq}[1]
           {\small {\tt #1} \normalsize}

\newcommand{\calD}{\mathcal D}
\newcommand{\calM}{\mathcal M} 
\newcommand{\calR}{\mathcal R} 
\newcommand{\calP}{\mathcal P} 
\newcommand{\calO}{O} 
\newcommand{\calT}{\mathcal T} 

\newcommand{\plan}{p} 

\newcommand{\Pre}{\textrm{PRE}}
\newcommand{\Add}{\textrm{ADD}}
\newcommand{\Del}{\textrm{DEL}}

\def\t{\mathcal{T}}

\def\m{\mathcal{M}}

\def\mlp{\tt AMDN}
\def\arms{\tt ARMS}
\def\aman{\tt AMAN}
\newcommand{\ignore}[1]{{}}

\begin{document}
\title{Learning Action Models from Plan Traces with Disordered Actions, Parallel Actions, Noisy States}

\author{Hankz Hankui Zhuo$^1$, Jing Peng$^1$, Subbarao Kambhampati$^2$
\\
$^1$School of Data and Computer Science, Sun Yat-Sen University, Guangzhou, China \\
$^2$Department of Computer Science and Engineering, Arizona State University, Tempe, Arizona, US \\
zhuohank@mail.sysu.edu.cn, pengj69@mail2.sysu.edu.cn, rao@asu.edu
}

\maketitle

\begin{abstract}
There is increasing awareness in the planning community 
that the burden of specifying complete domain models is
too high, which impedes the applicability of planning 
technology in many real-world domains. Although there
have many learning systems that help automatically learning
domain models, 
most existing work assumes that the input traces
are completely correct. A more realistic situation is
that the plan traces are disordered and noisy, such as plan traces described by natural language. 
In this paper we propose and evaluate an approach for
doing this. Our approach takes as input a set of plan traces with disordered actions and noise and outputs action models that can best explain the plan traces.
We use a MAX-SAT framework for learning, where the
constraints are derived from the given plan traces. 
Unlike traditional action models learners, the states in plan traces 
can be partially observable and noisy 
as well as the actions in plan traces can be disordered and parallel.
We demonstrate the effectiveness of our approach through a systematic
empirical evaluation with both IPC domains and the real-world dataset extracted from natural language documents.
\end{abstract}

\section{Introduction}
Most work in planning assumes that complete domain models are given as input in order to synthesize plans. However, there is increasing awareness that  building domain models at any level of completeness presents steep challenges for domain creators. As planning issues become more and more realistic, the action model will become more and more complex. So it is necessary for us to consider how to automatically attain the domain action model. Indeed, recent work in web-service composition (c.f. \cite{journal/aij/Bertoli10,conf/aaai/Hoffmann07}) and work-flow management (c.f. \cite{journal/is/Blythe04}) suggest that dependence on complete models can well be the real bottle-neck
inhibiting applications of current planning technology.

Attempts have been made to design systems to automatically learn domain models from pre-specified (or pre-collected) plan traces (i.e., each plan trace is composed of an action sequence with partial states between actions). For example, Amir \cite{DBLP:conf/ijcai/Amir05} presented a tractable and exact technique for learning action models known as Simultaneous Learning 
and Filtering (SLAF). Yang et al. \cite{DBLP:journals/ai/YangWJ07} proposed to learn STRIPS action models \cite{DBLP:journals/ai/FikesN71} from plan traces with partially observed states. Bryce et al. propose an approach called {\tt Marshal} to issue queries, and learns models by observing query answers, plan solutions, and direct changes to the model \cite{DBLP:conf/ijcai/BryceBB16}. \cite{DBLP:conf/aips/AinetoJO18} propose to learn STRIPS models via classical planning. These systems, however, are all based on the assumption that actions in plan traces are correct and totally ordered. 

In many real-world applications, however, plan traces are often extracted or built from raw data, such as monitoring signals, by off-the-shelf systems, due to the high cost of collecting (structured) plan traces by hand. Those plan traces are often with \textbf{disordered} and \textbf{parallel} actions, and \textbf{noisy} states. 
\ignore{For example, when experts meet in a meeting to decide on a planning issue, they usually propose some actions and describe the state after these actions are executed. During this process, experts often complement and modify each other's programs, which results in disordered actions traces.} 
\ignore{For example, the text ``before you go into the bus station, you should buy a ticket. Meanwhile please remember to bring the luggage with you'' addresses the procedure of go traveling. The action sequences that can be extracted by off-the-shelf approaches such as \cite{DBLP:conf/ijcai/FenceZK18,DBLP:conf/acl/BranavanKLB12} is ``goto(station), buy(ticket), bring(luggage)'', where ``buy(ticket)'' and ``bring(luggage)'' should be executed before ``goto(station)'', suggesting they are disordered, and ``buy(ticket)'' and ``bring(luggage)'' can be executed in parallel, i.e., they are parallel actions. States generated from text can be noisy as well due to the challenges of natural language comprehension for extracting states.} \emph{For example, network monitoring\footnote{https://www.dnsstuff.com/network-monitoring-software} is a system used to constantly monitor a computer network for slow or failing components and possible attacks. The ``actions'' (e.g., notifications of failures or attacks), which are collected by the monitoring system, are often disordered due to network traffic or other unexpected factors. There are also parallel actions collected due to many uncorrelated events that happen in parallel from the network. The collected states are often noisy as well due to possible errors introduced by the monitoring system.}

\ignore{
Moreover, more than one action may be executed at the same time, i.e., parallel actions are pervasive \cite{DBLP:conf/aaai/AghighiB17}. }

There have indeed been approaches that consider the issues of noisiness in plan traces. For example, Mourao et al. propose to learn action models from noisy observations of intermediate states, they assume actions are totally ordered and correct \cite{DBLP:conf/uai/MouraoZPS12}. Zhuo and Kambhampati designed a novel approach called {\tt AMAN} based on graphical model to consider actions being noisy in plan traces \cite{DBLP:conf/ijcai/ZhuoK13}. They, however, do not allow actions to be disordered (or parallel), or states to be noisy.  
\ignore{
They, however, assume that noisy actions are similar to their corresponding correct actions when building the conditional distribution of correct actions given noisy actions,  }Recent approaches \cite{DBLP:conf/aaai/AsaiF18,DBLP:journals/jair/KonidarisKL18,DBLP:conf/aips/AsaiK19,DBLP:journals/corr/abs-1905-12006} aim to learn state representations from raw data such as images, which can be noisy, to help high-level planning. They, however, assume action sequences are correct, i.e., without disordered and parallel actions. 

In this paper, we aim to learn action models from plan traces with disordered actions, parallel actions, and noisy states. This is challenging in the sense that each of the three uncertain cases can harm the learning quality of action models. To address the challenge, we build three types of constraints from plan traces, namely \emph{disorder constraints}, \emph{parallel constraints}, and \emph{noise constraints}, to capture information from disordered actions, parallel actions, and noisy states, respectively. We then solve the constraints with an off-the-shelf weighted MAX-SAT solver, such as \cite{DBLP:conf/ijcai/BacchusHJS18}, and convert the solution to action models. We denote our approach by {\mlp}, which stands for Learning \textbf{A}ction \textbf{M}odels from plan traces with \textbf{D}isordered and parallel actions and \textbf{N}oisy states. We will evaluate {\mlp} on both IPC\footnote{http://www.icaps-conference.org/index.php/Main/Competitions} domains and a real-world dataset extracted from natural language documents.  

Although our {\mlp} approach uses the same MAX-SAT framework with previous work such as ARMS \cite{DBLP:journals/ai/YangWJ07} and ML-CBP \cite{DBLP:conf/aaai/ZhuoNK13,DBLP:journals/ai/ZhuoK17}, the constraints we built from plan traces with disordered and parallel actions, and noisy states are totally different from the ones built by previous work. Compared to building constraints based on perfectly correct plan traces as done by previous work, building new constraints based on disordered and parallel actions and noisy states is challenging due to complicated relationships among disordered and parallel actions and noisy states. Recent work, such as {\tt AMAN} \cite{DBLP:conf/ijcai/ZhuoK13}, tended to give up the MAX-SAT based framework (such as {\tt ARMS}) and turn to other different framework (such as {\tt AMAN}, the graphical model based approach) when there was noise involved. It is, however, not specific to disorder and parallelism, as is demonstrated in our experimental results. Our {\mlp} approach, i.e., based on the MAX-SAT framework, is much more natural and effective by building new constraints regarding disordered actions, parallel actions and noisy states. 
\ignore{
Note that we consider learning action models from structured plan traces (which are composed of action sequences with states in between) instead of unstructured data in the form of natural language. There have been many approaches to extracting actions from texts in natural language processing community such as \cite{DBLP:conf/acl/BranavanKLB12}, which is out of the scope of this paper.}

\ignore{We use the MAX-SAT framework for learning, where the constraints are
derived from the executability of the given plan traces. Unlike traditional
action models learners which demand high quality plan traces, our entered 
plan traces is disordered and noisy. 
First of all, we build \emph{plan constraints} form the executed order
of actions in the entered plan traces. In addition, our algorithm build
\emph{state constraints} by using information observed from state in the 
entered plan traces. Finally, we build \emph{action constraints}, which 
are imposed on individual actions. Since the entered plan traces are 
disordered and noisy, we give each constraint a corresponding weight.
These weight are obtained by calculating the probability of this 
constraint being established as well as considering the sensitivity
of specific domain to disorder and noise.
The higher the weight, the greater the degree of confidence we hold in 
this constraint and the higher its priority in the MAX-SAT solver.
We demonstrate the effectiveness of our approach, 
called {\mlp} which stands for Learn \textbf{A}ction 
\textbf{M}odels from \textbf{D}isordered and \textbf{N}oisy plan traces,
and present a systematic empirical evaluation that demonstrates . }

In the remainder of the paper, we first review previous work related to our approach, and
then present the formal definition of our problem. After that, we provide
the detailed description of our {\mlp} algorithm and evaluate
{\mlp} in two planning domains and a real-world dataset. Finally we conclude our paper with future work.

\section{Related Work}
There have been many approaches on learning action models from plan traces. Previous
research efforts differ mainly on whether plan traces consist of actions
and intermediate states (c.f. \cite{DBLP:conf/icml/Gil94}) or 
only actions (c.f. \cite{DBLP:journals/ai/YangWJ07,DBLP:journals/ai/ZhuoYHL10,DBLP:conf/aips/ZhuoYPL11,DBLP:journals/ai/Zhuo014,DBLP:journals/ai/ZhuoM014,DBLP:conf/aaai/Zhuo15}).
While the latter assumption makes the problem harder than the former, in both cases,
whatever observed is assumed to be observed perfectly. Both of them assume
non-noisy plan observations. While there has been previous work on learning
probabilistic action models(e.g. \cite{DBLP:journals/jair/PasulaZK07} and
\cite{DBLP:conf/aaai/ZettlemoyerPK05}), they also assume non-noisy plan observations. Recently, Gregory et al. present an algorithm called LOP to induce static predicates to the learning system, which finds a set of minimal static predicates for each operator that preserves the length of the optimal plan \cite{DBLP:conf/ijcai/GregoryC16}. Instead of an action model, a set of successfully executed plans are given and the task is to generate a plan to achieve the goal without failing \cite{ijcai2017-615}. Lindsay et al. propose to learn action models action descriptions in the form of restricted template \cite{DBLP:conf/aips/LindsayRFHPG17}. To further relax the correctness requirement of action sequences, Aineto et al. \cite{DBLP:journals/ai/Aineto19} propose to learn action models based on partially observed actions in plan traces. Despite the success of those approaches, they all assume observed actions are correctly ordered and states are not noisy. 

\ignore{
Latplan \cite{DBLP:conf/aaai/AsaiF18} connected a subsymbolic neural network (NN) system and a symbolic Classical Planning system to solve various visually presented puzzle domains. The State AutoEncoder  (SAE)  neural  network  in  Latplan  generates  a  set  of propositional symbols from the training images with no additional  information  and  provides  a  bidirectional  mapping between images and propositional states. The system then solves the propositional planning problem using a classical planner Fast Downward \cite{DBLP:conf/aips/Helmert04} and returns an image sequence that solves the puzzle by decoding the intermediate propositional states of the plan. 
Another  recent  approach replacing SAE/AMA2with InfoGAN \cite{NIPS2018_8090} has no explicit mechanism for improving  the  stability  of  the  binary representation.

\cite{DBLP:journals/jair/KonidarisKL18} generates a symbolic state space from the low-level sensor inputs but requires the high-level action symbols. Our approach complements the usability of these approaches by providing more stable symbols. Previous  work  in  Learning  from  Observation  \cite{DBLP:conf/icra/BarbuNS10,DBLP:conf/aaai/Kaiser12},  which could produce propositions from the observations (unstructured input) with the help of hand-coded symbol extractors (e.g.ellipse detectors for tic-tac-toe), typically assume a perfect indoor environment with little noisy interference, there-fore do not have to address the stability of the symbols.
}
\section{Preliminaries and Problem Definition}
A complete STRIPS domain can be defined by a tuple $\calD = \langle \calR, \calM \rangle$, where $\calR$ is a set of predicates with typed objects and $\calM$ is a set of action models. Each action model is a quadruple $\langle a,\Pre(a), \Add(a), \Del(a) \rangle$, where $a$ is an action name with zero or more parameters, $\Pre(a)$ is a precondition list specifying the conditions under which $a$ can be applied, $\Add(a)$ is an adding list and $\Del(a)$ is a deleting list indicating the effects of $a$. 
We denote $\calR_{\calO}$ as the set of propositions instantiated from $\calR$ with respect to a set of typed objects $\calO$. Given $\calD$ and $\calO$, we define a planning problem as $\calP = \langle \calD, s_0, g \rangle$, where $s_0 \subseteq \calR_{\calO}$ is an initial state, $g \subseteq \calR_{\calO}$ are goal propositions.

We denote a set of \emph{parallel actions} by $\Psi$ in a plan, where actions in $\Psi$ can be executed in any order or simultaneously. For example, let $\Psi$ be $\{a_1,a_2\}$. The sequence $\langle s_0, \Psi, s_1 \rangle$ is equivalent to $\langle s_0, a_1, a_2,s_1 \rangle$ or $\langle 
s_0, a_2, a_1, s_1 \rangle$. A solution plan to $\calP$ with respect to model
$\calD$  is a sequence of \emph{parallel actions} $\plan = 
\langle \Psi_1, \Psi_2, \ldots, \Psi_n \rangle$ that achieves 
goal $g$ starting from $s_0$. If actions $a_x \in \Psi_i$ and $a_y \in \Psi_j$, 
we define the \emph{distance} of $a_x$ and $a_y$ by $|a_x-a_y|=|i-j|$. If the positions of $a_x$ and $a_y$ are exchanged, we say they are \emph{disordered} with respect to their \emph{correct} order in the plan.

A plan trace $t$ is defined by $t=\langle s_0, \Psi_1, s_1, ..., \Psi_n, 
g \rangle$, where actions can be disordered. We assume that the probability of two actions being disordered 
decreases as their \emph{distance} increases. This is reasonable in the sense that actions in a short-term horizon are more likely to be disordered than a long-term horizon.  State $s_i$ is both partial and noisy, indicating some propositions are missing in $s_i$ and some propositions in $s_i$ are incorrect.  We denote a set of plan traces as $\calT$. Our problem can be 
defined by: given as input a set of observed plan traces 
$\calT$, our approach outputs a domain model $\calM$ that
best explains the observed plan traces. 

\begin{figure}[!ht]
	\centering
	\includegraphics[width=0.48\textwidth]{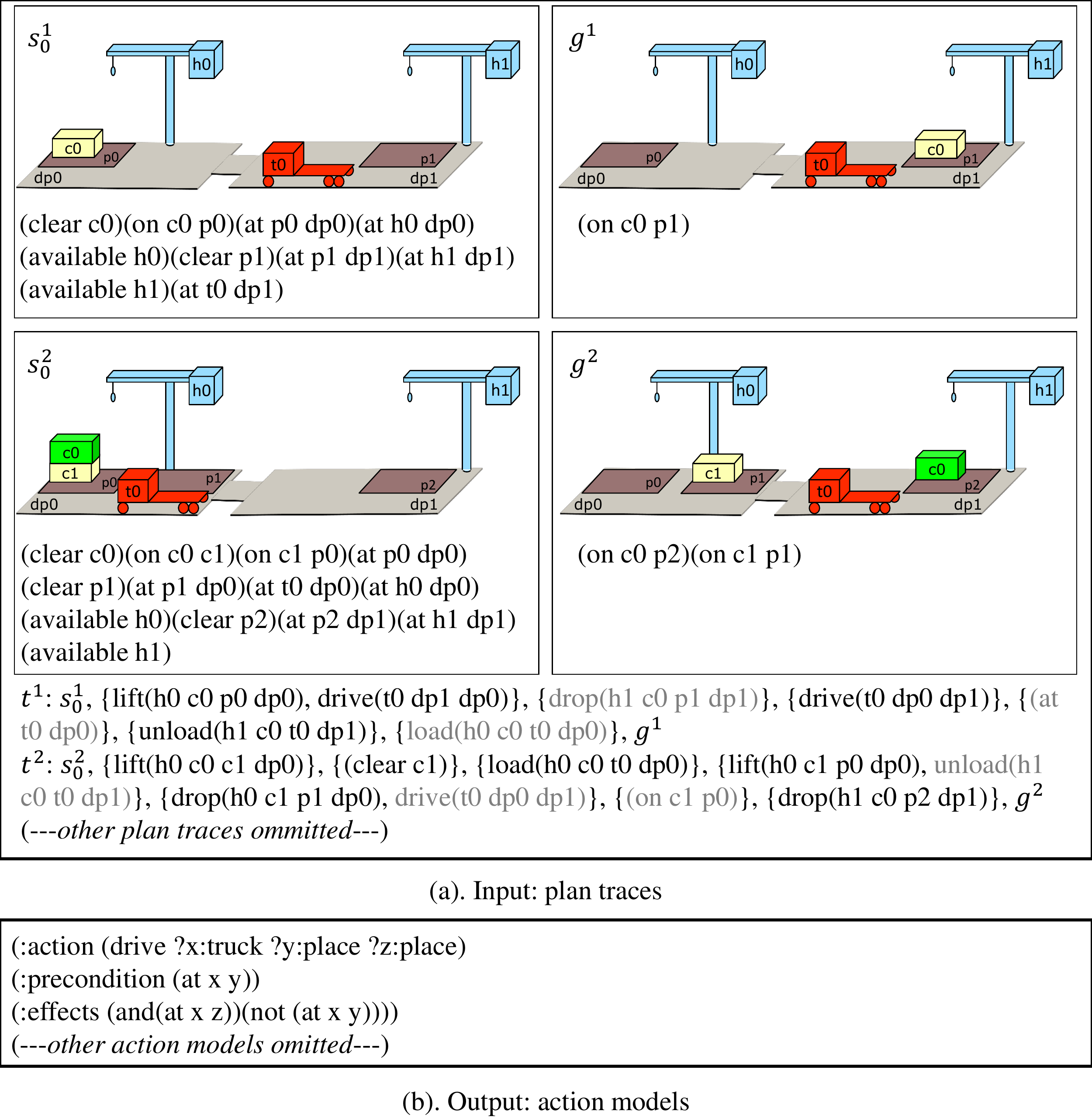}
	\caption{An example of our learning problem.}
	\mvp
	\label{figure:ipop}
\end{figure}

An example of our learning problem for the \emph{depots}\footnote{http://planning.cis.strath.ac.uk/competition/}
domain can be found in Figure
\ref{figure:ipop}, which is composed of two parts:
plan traces as input (Figure \ref{figure:ipop}(a)) and 
action models as output (Figure \ref{figure:ipop}(b)). 
In Figure \ref{figure:ipop}(a), $t^1$ and $t^2$ are two 
plan traces, where initial states and goals are drawn
over. The dark parts indicate the incorrect propositions or
disorder actions. For example, in $t^1$, ``drop(h1 c0 p1 dp1)" and ``load(h0 c0 
t0 dp0)" are disordered, ``(at t0 dp0)" is a noisy proposition which should be ``(at t0 dp1)". In Figure \ref{figure:ipop}
(b), the action model ``drive'' is one of the output action models of our algorithm.

\ignore{
\settowidth\rotheadsize{\theadfont ----other plan traces ommitted--------}
\begin{table}[h]
	\centering
	\caption{An example of our learning problem.}
	\begin{tabular}{|c|c|c|}
		\hline 
		\multicolumn{3}{|c|}{Input:}\\
		\hline	
		\makecell{$s_0^1$}&\makecell{$s_0^2$}&\multirow{11}*{\rothead{----other plan traces ommitted----}} \\
		\cline{1-2}
		\makecell{lift(h0\ c0\ p0\ ds0),\\
			drive(t0\ dp1\ dp0)}&\makecell{lift(h0\ c1\ c0\ dp0)} &~\\
		\cline{1-2}
		&&~ \\
		\cline{1-2}
		\makecell{\textcolor[rgb]{0.4,0.4,0.4}{\emph{drop(h1\ c0\ p1\ dp1)}}}&\makecell{load(h0\ c1\ t0\ dp0)} &~\\
		\cline{1-2}
		&&~ \\
		\cline{1-2}
		\makecell{drive(t0\ dp0\ dp1)}&\makecell{lift(h0\ c0\ p0\ dp0),\\
			 \textcolor[rgb]{0.4,0.4,0.4}{\emph{unload(h1\ c1\ t0\ dp1)}}} &~\\
		\cline{1-2}
		\textcolor[rgb]{0.4,0.4,0.4}{\emph{(at t0 dp0)}}&&~ \\
		\cline{1-2}
		\makecell{unload(h1\ c0\ t0\ dp1)}&\makecell{drop(h0\ c0\ p1\ dp0),\\
				\textcolor[rgb]{0.4,0.4,0.4}{\emph{drive(t0\ dp0\ dp1)}}} &~\\
		\cline{1-2}
		&\textcolor[rgb]{0.4,0.4,0.4}{\emph{(on c0 p0)}}&~ \\
		\cline{1-2}
		\makecell{\textcolor[rgb]{0.4,0.4,0.4}{\emph{load(h0\ c0 \ t0\ dp0)}}}&\makecell{drop(h1\ c0\ p2\ dp1)} &~\\
	    \cline{1-2}
		$g^1$&$g^2$&~ \\
		\hline
		\multicolumn{3}{|c|}{Output:}\\
		\hline
		\multicolumn{3}{|c|}{\makecell[l]{(:action(drive ?x:truck ?y:place ?z:place)\\
				(:precondition (at x y))\\
				(:effects (and(at x z))(not (at x y))))\\
			(---other action models omitted---)}} \\
		\hline
		\multicolumn{3}{c}{\makecell[l]{$s_0^1$:\{(clear c0)(on c0 p0)(at p0 dp0)
			(at h0 dp0)(available\\h0)(clear p1)
			(at p1 dp1)(at h1 dp1)(available h1)
			(at t0\\ dp1)\};\\$g^1$:\{(on c0 p1)\}.\\
		$s_0^2$:\{(clear c0)(on c0 c1)(on c1 p0)
		(at p0 dp0)(clear p1)\\(at p1 dp0)
		(at h0 dp0)(available h0)(at t0 dp0)
		(clear p2)\\(at p2 dp1)(at h1 dp1)
		(available h1)\};\\$g^2$:\{(on c0 p2)(on c1 p1)\}.}}
	\end{tabular}
	\label{tab:ipop}
\end{table}
}

\ignore{
An example input of our planning problem in \emph{blocks}\footnote{http://www.cs.toronto.edu/aips2000/} is shown in Figure \ref{example:input}, which is composed of three parts: incomplete action models (Figure \ref{example:input}(a)), initial state $s_0$ and goal $g$ (Figure \ref{example:input}(b)), and a plan example set (Figure \ref{example:input}(c)). In Figure \ref{example:input}(a), the dark parts indicate the missing predicates. In Figure \ref{example:input}(c), $p_1$ and $p_2$ are two plan traces, where initial states and goals are bracketed. An example output is a solution to the planning problem given in Figure \ref{example:input}, i.e., ``\seq{(unstack C A)(putdown C)(pickup B)(stack B A)(pickup C)(stack C B)(pickup D)(stack D C)}''.
\begin{figure}[!ht]
  \centering
  \includegraphics[width=0.37\textwidth]{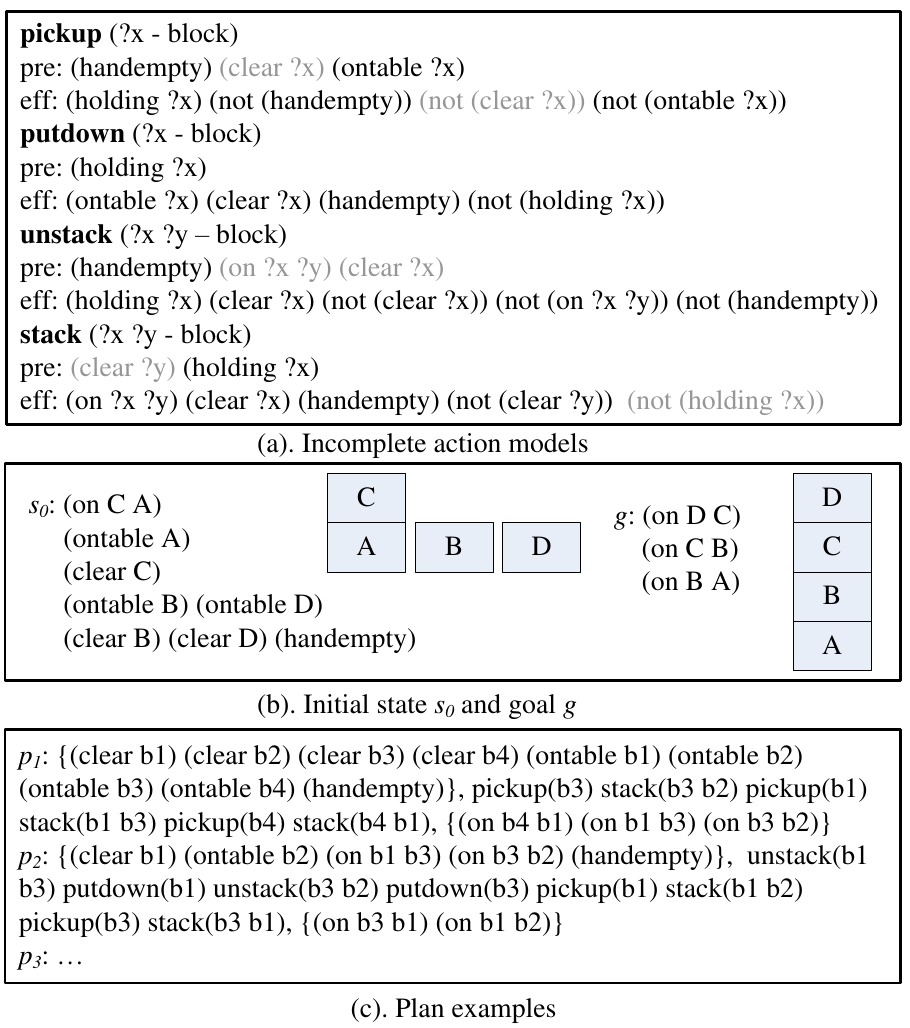}
  \caption{An input example of {\mlp} for the \emph{blocks} domain}
\mvp\mvp
\label{example:input}
\end{figure}
}
\section{The {\mlp} Approach}
In this section we present {\mlp} approach in detail. An overview of our approach is shown in Algorithm \ref{algorithm:main}. We first build sets of \emph{disorder constraints}, \emph{parallel constraints} and \emph{noise constraints}, to encode the information of \emph{disorder, parallelism, and noise}. After that we solve these constraints with an off-the-shelf MAX-SAT solver, and then convert the solution to action models.
\begin{algorithm}[!ht]
\caption{An overview of our {\mlp} approach}\label{algorithm:main}
\textbf{Input:} a set of plan traces $\t$. \\
\textbf{Output:} a set of action models $\m$.

\begin{algorithmic}[1]
	\STATE build \emph{disorder constraints};
	\STATE build \emph{parallel constraints};
	\STATE build \emph{noise constraints};
	\STATE solve all constraints with a MAX-SAT solver
	\STATE convert the solution to action models $\m$;
	\STATE return $\m$;
\end{algorithmic}
\end{algorithm}

\subsection{Building \emph{disorder constraints}}
In Step 1 of Algorithm \ref{algorithm:main} we aim to build a set of \emph{disorder constraints} to capture the information of disorder of actions in plan traces according to the correctness of executing procedure of plan traces. We denote this set of \emph{disorder constraints} by DC, which will be addressed in detail below.
\paragraph{Constraint DC}
Consider two adjacent sets of \emph{parallel actions} $\langle \Psi_i,\Psi_{i+1} 
\rangle$ in a plan trace. We assume that any action $a\in\Psi_{i+1}$ is not parallel with any action in $\Psi_i$ (otherwise $a$ will be put in $\Psi_i$). In other words, for each action $a_y\in\Psi_{i+1}$, there exists $a_x\in\Psi_i$, such that there are interactions between $a_y$ and $a_x$, which are as shown below: 
\begin{enumerate}
	\item 
	A \emph{proposition} $r$ in the precondition list of $a_x$ but not in
	the \emph{delete} list of $a_x$ will be deleted by $a_y$, which can be encoded
	by: 
	$\exists r \  r\in \textrm{PRE}(a_x) \wedge r\not\in\textrm{DEL}(a_x)
	\wedge r\in\textrm{DEL}(a_y).$
	\item 
	A \emph{proposition} $r$ in the precondition list of $a_y$ is added
	by $a_x$, which can be encoded by: 
	$\exists r \  r\in\textrm{ADD}(a_x) \wedge r\in\textrm{PRE}(a_y).$
	\item 
	A \emph{proposition} $r$ is in the add list of $a_x$ and it is 
	in the delete list of $a_y$, which can be encoded by:
	$\exists r \  r\in\textrm{ADD}(a_x)\wedge r\in\textrm{DEL}(a_y).$
	\item 
	A \emph{proposition} $r$ is in the delete list of $a_x$ and it is 
	in the add list of $a_y$, which can be encoded by: 
	$\exists r \  r\in\textrm{DEL}(a_x)\wedge r\in\textrm{ADD}(a_y).$
	\end{enumerate}
That is to say, the following constraint holds if $a_x$ and $a_y$ are ordered:
\begin{eqnarray}\label{dc1}
&(r\in\textrm{PRE}(a_x)\wedge r\not\in\textrm{DEL}(a_x) 
\wedge r\in\textrm{DEL}(a_y)) \notag\\
&\vee (r\in\textrm{ADD}(a_x)\wedge r\in\textrm{PRE}(a_y)) \notag\\
&\vee (r\in\textrm{ADD}(a_x)\wedge r\in\textrm{DEL}(a_y)) \notag\\
&\vee (r\in\textrm{DEL}(a_x)\wedge r\in\textrm{ADD}(a_y)). 
\end{eqnarray}
If $a_x$ and $a_y$ are disordered, we need to swap the positions of $a_x$ and $a_y$ and the resulting constraint should hold correspondingly: 
\begin{eqnarray}\label{dc2}
&(r\in\textrm{PRE}(a_y) \wedge r\not\in\textrm{DEL}(a_y) 
\wedge r\in\textrm{DEL}(a_x)) \notag\\
&\vee (r\in\textrm{ADD}(a_y)\wedge r\in\textrm{PRE}(a_x)) \notag\\
&\vee (r\in\textrm{ADD}(a_y)\wedge r\in\textrm{DEL}(a_x)) \notag\\
&\vee (r\in\textrm{DEL}(a_y)\wedge r\in\textrm{ADD}(a_x)).
\end{eqnarray}
We assume the prior probability of two actions being disordered, denoted by $p(a_x,a_y)$, depends on a set of features $\textbf{f}(a_x,a_y)$, which gives us the flexibility to incorporate a diverse range of features \cite{DBLP:journals/pami/PietraPL97}. $\textbf{f}(a_x,a_y)=(f_1,f_2,\ldots,f_k)$ is an $k$-dimensional feature representation. The distribution over all action pairs 
\begin{equation}\label{dcw}
p(a_x,a_y)=\frac{\exp(\Vec{\theta}\cdot\textbf{f}(a_x,a_y))}{\sum_{a'_x\in A,a'_y\in A,a'_x\neq a'_y}\exp(\Vec{\theta}\cdot\textbf{f}(a'_x,a'_y))}
\end{equation}
where $\Vec{\theta}=(\theta_1,\theta_2,\ldots,\theta_k)$ is a $k$-dimensional vector of parameters that can be learnt from labeled training data. In this paper, we do not assume we have such training data for learning $\Vec{\theta}$. We thus empirically set $\theta_i$ to be $\frac{1}{k}$ for all $1\leq i\leq n$, viewing each feature $f_i$ is equally important. We used 40 features to specify the probability of disorder between two actions. The following are example features we used in our experiment, which is possibly extended to more features in future study: 
\begin{itemize}
    \item $f_1$: the number of objects shared by actions $a_x$ and $a_y$. 
    \item $f_2$: if the numbers of parameters of actions $a_x$ and $a_y$ are identical, the value of $f_2$ is one; otherwise it is zero. 
    \item $f_3-f_{10}$: these features are extracted to describe the similarity between $a_x$ and $a_y$ regarding their semantics in real-world applications, such as action directions.
    \item $f_{11}-f_{40}$: these features are built to describe the similarity between $a_x$ and $a_y$ based on text descriptions of action semantics from web search \cite{DBLP:journals/ai/Zhuo014} (a text description is represented by a 30-dimensional vector using the document-representation approach \cite{DBLP:conf/icml/LeM14}.
\end{itemize}
Based on Equation (\ref{dcw}), we can calculate the probability $p(a_x,a_y)$ for Constraint (\ref{dc2}) and the probability $1-p(a_x,a_y)$ for Constraint (\ref{dc1}). To assign a weight to Constraints (\ref{dc1}) and (\ref{dc2}) in the MAX-SAT framework, we multiply each probability by a maximal weight (denoted by $w_{max}$). That is to say, the weights of Constraints (\ref{dc1}) and (\ref{dc2}) are $(1-p(a_x,a_y))\times w_{max}$ and $p(a_x,a_y)\times w_{max}$, respectively.
\ignore{
If the actions are ordered, DC can
be hard constraints. When the actions are disordered, we set
the weight as follows:
$w_d=w_{do}P_d,$
where $w_{do}$ is the original weight (we empirically set $w_{do}$ to be 5)
and $P_d$ is the belief of this constraint. In this paper, 
we calculate $P_d$ by  
\begin{equation}\nonumber
	P_d=
	\begin{cases}
	\frac{\theta_{diso}}{K_1}& K_1\neq0\\
	1& K_1=0
	\end{cases}
\end{equation}
where $\theta_{diso}$ is the probability of disorder.
$K_1(a_i, a_j)$ is the \emph{distance} between $\Psi_j$ 
and $\Psi_{i+1}$, which can be defined by $K_1(a_i, a_j)=|i+1-j|$. 
For instance, if $a_{i1},a_{i2} \in \Psi_i, a_j \in \Psi_{i+1}$ and $a_k \in \Psi_{i+2}$, where
$\Psi_i$, $\Psi_{i+1}$ and $\Psi_{i+2}$ are adjacent sets of \emph{parallel actions} 
in a plan trace $t$, 
we have $K_1(a_{i1}, a_{i2})=1$, $K_1(a_{i1}, a_j)=0$ and $K_1(a_{i1}, a_k)=1$.
}

\subsection{Building \emph{parallel constraints}}
In Step 2 of Algorithm \ref{algorithm:main}, we build a set of \emph{parallel constraints}.
To make sure that the learned action models are succinct and consistent with the
STRIPS language,
we first enforce a set of hard constraints that must be
satisfied by action models, which were also built by \cite{DBLP:journals/ai/YangWJ07}. We then build a set of soft constraints between actions 
in the same set of \emph{parallel actions}.
We formulate the constraints as follows and denote them by PC.

An action may not add a \emph{proposition} which
already exists before the action is applied. We formulate the constraints
as follows and denote them by P.1:
$r\in\textrm{ADD}(a)\Rightarrow r\not\in\textrm{PRE}(a).$

An action can not delete a \emph{proposition} which does not exist
before the action is applied. We formulate the constraints
as follows:
$r\in\textrm{DEL}(a)\Rightarrow r\in\textrm{PRE}(a).$
To make sure these constraints are hard, we assign these constraints with the maximal weight $w_{max}$. 

Consider two action $a_x\in\Psi_i$ and $a_y\in\Psi_{i+1}$. If any $a'_x\in\Psi_i$ adds or deletes a \emph{proposition} 
$r$, $r$ cannot be in the add list or delete list of $a_x$. 
In other words, the following constraint holds:
\begin{eqnarray}\label{pc1}
\textrm{ADD}(a'_x)  \cap  \textrm{DEL}(a'_x) \cap \textrm{ADD}(a_x) 
\cap \textrm{DEL}(a_x)=\varnothing.
\end{eqnarray}
If $a_x$ and $a_y$ is disordered, i.e., $a_y$ should be in $\Psi_i$ and $a_x$ should be replaced by $a_y$ in Constraint (\ref{pc1}), we have:
\begin{eqnarray}\label{pc2}
\textrm{ADD}(a'_x)  \cap  \textrm{DEL}(a'_x) \cap \textrm{ADD}(a_y) 
\cap \textrm{DEL}(a_y)=\varnothing.
\end{eqnarray}
Similar to DC, the weights of Constraints (\ref{pc1}) and (\ref{pc2}) are $(1-p(a_x,a_y))\times w_{max}$ and $p(a_x,a_y)\times w_{max}$, respectively. 
\ignore{the actions are ordered, P.3 can be hard constraints. When the actions are disordered, we set the weight as follows:
$w_p=w_{po}P_p,$
where $w_{po}$ is the original weight (similar to $w_{do}$, we empirically set $w_{po}$ to be 5)
and $P_p$ is the belief of this constraint. In this paper, 
we calculate $P_p$ by  
\begin{equation}\nonumber
P_p=
\begin{cases}
\frac{\theta_{diso}}{K_2}& K_2\neq0\\
1& K_2=0
\end{cases}
\end{equation}
where $K_2(a_i, a_j)$ is the \emph{distance} between $\Psi_j$ 
and $\Psi_i$, which can be defined by $K_2(a_i, a_j)=|i-j|$. 
For instance, if $a_{i1},a_{i2} \in \Psi_i, a_j \in \Psi_{i+1}$, and $a_k \in \Psi_{i+2}$, where
$\Psi_i$, $\Psi_{i+1}$ and $\Psi_{i+2}$ are adjacent sets of \emph{parallel actions} 
in a plan trace $t$, 
we have $K_2(a_{i1}, a_{i2})=0$, $K_2(a_{i1}, a_j)=1$ and $K_2(a_{i1}, a_k)=2$.
}

\subsection{Building \emph{noise constraints}}
In Step 3 of Algorithm \ref{algorithm:main}, we build a set of \emph{noise constraints} to encode the information from noisy states in plan traces. We denote noise constraints by NC.

Consider a pair $\langle a,r \rangle $, where $a$ is an action 
and $r$ is a \emph{proposition}. If the number of its occurrence over all of the plan traces is higher than the 
threshold $\delta$, $r$ is viewed as a correct proposition (i.e., not a noisy proposition) and should not be deleted by $a$ since $r$ exists after $a$. That is to say, the following constraint holds:

\begin{equation}\label{n1}
\langle a,r \rangle \Rightarrow r \not\in \textrm{DEL}(a).
\end{equation}
Consider a sequence 
$\langle s_0,a_0,a_1,...,a_n,r \rangle$, where
$a_i$ $(0\le i \le n)$ is an action and
$r\not\in s_0$ is a \emph{proposition}. $r$ must be in the add list of some
$a_i$ $(0\le i \le n)$. That is to say, the following constraint holds:
\begin{eqnarray}\label{n2}
r \in \textrm{ADD}(a_0) \cup \textrm{ADD}(a_1) \cup ...
\cup \textrm{ADD}(a_n).
\end{eqnarray}

Consider a pair $\langle r, a \rangle$, where $r$ is a \emph{proposition} appearing before action 
$a$. If the number of its occurrence over all of the plan traces is higher than the 
threshold $\delta$, $r$ is viewed as a correct proposition and a precondition of $a$, i.e., 
\begin{equation}\label{n3}
    r\in \textrm{PRE}(a).
\end{equation}

The weights of Constraints (\ref{n1})-(\ref{n3}) are calculated by the product of the ratio of occurrences of $r$ over all propositions and $w_{max}$, i.e., \[
\frac{\textrm{occurrences of }r}{\textrm{occurrences of all propositions}}\times w_{max}.\]
\ignore{
where the $\theta(\langle r_0, a_0 \rangle)=\frac{\textrm{number\ of}\ 
\langle r_0, a_0 \rangle}{\textrm{number\ of\ all}\ \langle r, a_0 \rangle}$.
}
\ignore{
These constraints are designed to be soft, since the
observed states are noisy and the actions are disordered. Even without
noise, we can not be certain that N.3 is correct. 
So the weight of these constraints is dynamic, which are designed to
be 
$w_n=w_{no}P_nP_f$,
where $w_{no}$ is the original weight (similar to $w_{do}$, we empirically set $w_{no}$ to be 5), 
$P_n$ is the belief of the constraint and $P_f$ is the frequency of
observation.
In N.1, we calcualte $P_n$ by using the function $P_n=\frac{\theta_{diso}}{K_3}$ if $K_3\neq0$, and $P_n=1$ otherwise.
$K_3(a_i, r)$ is the \emph{distance} between $\Psi_i$ 
and the first \emph{parallel actions} before $r$, where $a_i\in\Psi_i$.
So $K_3(a_i, r)=|i-j|$, where $a_i\in \Psi_i$, $r\in s$ and $\Psi_j$ is the first set of \emph{parallel actions} before $s$.  
For instance, if $a_i \in \Psi_i$, $a_j \in \Psi_{i+1}$, $r\in s_i$ and $\langle \Psi_i, s_i,\Psi_{i+1} \rangle$, we have $K_3(a_i, r)=0$ and $K_3(a_j, r)=1$.
In N.2, $P_n$ is defined by
\[
P_n=
\frac{\textrm{the\ number\ of\ actions\ executed\ before}\ r}{\textrm{the\
	total\ number\ of\ actions\ in\ this\ trace}}.
\]
In N.3, we calcualte $P_n$ using the function $P_n=\frac{\theta_{diso}}{K_4}$ if $K_4\neq0$, $P_n=1$ otherwise.
$K_4(r, a_i)$ is the \emph{distance} between $\Psi_i$ 
and the first set of \emph{parallel actions} after $r$, where $a_i\in\Psi_i$.
So $K_4(r, a_i)=|i-j|$, where $a_i\in \Psi_i$, $r\in s$ and $\langle s, \Psi_j \rangle$.  
For instance, if $a_i \in \Psi_i$, $a_j \in \Psi_{i+1}$, $r\in s_i$ and $\langle  s_i,\Psi_i,\Psi_{i+1} \rangle$, we have $K_4(r, a_i)=0$ and $K_4(r, a_j)=1$.
$P_f$ is derived from the statistics. For
instance, in N.1, if we observed $\langle a, r_1 \rangle$ one time and 
observed $\langle a, r_2 \rangle$ nine times, $P_f(\langle
a, r_1 \rangle)=0.1$ and $P_f(\langle a, r_2\rangle)=0.9$. This is 
similar in N.2 and N.3.
}
\subsection{Solving Constraints}
In Steps 4 and 5 of Algorithm \ref{algorithm:main}, we solve all of the weighted constraints and convert the solution to action models.
We put all constraints together and solve them with a weighted MAX-SAT
solver \cite{journal/jair/li07}. The greater the weight of the constraint,
the higher its priority in the MAX-SAT solver. We exploit MaxSatz
\cite{journal/jair/li07} to solve all the constraints, and
attain a \emph{true} or \emph{false} assignment to maximally satisfy
the weighted constraints. Given the solution assignment, the
construction of action models $\m$ is
straightforward, e.g., if ``$r\in \textrm{ADD}(a)$'' is assigned
\emph{true} in the result of the solver, $r$ will be converted into an
effect of $a$.

\section{Experiments}
In this section we evaluate our approach with comparison to state-of-the-art approaches. We will first introduce the domains in which we conducted our experiments and the criterion we used to measure our approach as well as baselines. After that we present our experimental results with respect to various aspects. 
\subsection{Domains}
We evaluate our {\mlp} approach in three planning domains, i.e., 
\emph{blocks}\footnote{http://www.cs.toronto.edu/aips2000/},
\emph{driverlog}$^4$, and \emph{depots}$^3$, and a real-world domain, i.e., ``CookingTutorial''\footnote{http://cookingtutorials.com/}.
In the three planning domains, we generate 120  plan traces with disordered actions, parallel actions, and noisy states. \ignore{For each plan trace, the rate of observations
$\theta_1$ means that only $\theta_1$ of the \emph{propositions} in each state can be observed.
If the probability of noise is $\theta_2$, for each one of our observed
\emph{propositions}, the probability of being replaced by another different and random 
\emph{proposition} is $\theta_2$.
If the probability of disorder is $\theta_3$, for all $a_x\in\Psi_i$, 
$a_y\in\Psi_j$ and $\Psi_i,\Psi_j\in t$, the probability of exchanging the
order of $a_x$ and $a_y$ is $\frac{\theta_3}{K}$, where $K$ is the 
\emph{distance} between $a_x$ and $a_y$. In the experiments, } The generating process is as shown below:
\begin{enumerate}
	\item We first generated three sets of \emph{correct} plan traces using an off-the-shelf planner, FF\footnote{http://fai.cs.uni-saarland.de/hoffmann/ff.html}, to solve randomly generated planning problems with ground-truth models of domains \emph{blocks}, \emph{driverlog}, and \emph{depots}, respectively. Each set has 120 plan traces (actions are total order in each plan trace). The average lengths of domains \emph{blocks}, \emph{driverlog} and \emph{depots} are 65, 85 and 93, respectively.
		\item For each plan trace, we generate disordered actions by the following procedure: for each pair of sets of parallel actions $\Psi_i$ and $\Psi_j$, we exchange the order of two actions $a_i\in\Psi_i$ and $a_j\in\Psi_j$ with the probability $\frac{p}{d(\Psi_i,\Psi_j)}$, where $d(\Psi_i,\Psi_j)$ is the \emph{distance} between $\Psi_i$ and $\Psi_j$, and $p$ is the prior probability of two actions being disordered. 
	\item For each plan trace $t$, we built \emph{parallel actions} as follows: \begin{enumerate}
	    \item Let $k=1$, the set of parallel actions $\Psi_k=\emptyset$ and the state (after $\Psi_k$) $s'_k=\emptyset$.
	    \item For each action $a_i\in t$ ($1\leq i\leq |t|$), if the intersection between $a_i$'s conditions (including precondition, deleting and adding lists) and the conditions of each action in $\Psi_k$, let $\Psi_k=\Psi_k\cup\{a_i\}$ and state $s'_k=s'_k\cup s_i$, i.e., $a_i$ does not influence any action in $\Psi_k$ or is influenced by $\Psi_k$; otherwise, let $k=k+1$, $\Psi_k=\emptyset$ and $s'_k=\emptyset$.
	\end{enumerate}
	\ignore{
	Initialize an empty trace $t'=\langle \rangle$ and a \emph{parallel actions} $\Psi=\{a_0\}$. We denote by $s_{first}$, $s_{second}$ and $s_{third}$ the first, the second and the third state in the trace $t$, respectively. While there are actions in $t$, we pop the first action $a$ and test if $\langle s_{first}, \Psi, a, s_{third} \rangle$ is equivalent to $\langle s_{first}, a, \Psi, s_{third} \rangle$. If \emph{true}, we push $a$ into $\Psi$ and delete $s_{second}$. If \emph{false}, we push $s_{first}$ and $\Psi$ into $t'$, delete $s_{first}$ from $t$ and replace $\Psi$ by $\{a\}$. For instance, we have $t=\langle s_0,a_0,s_1,a_1,...,g\rangle$. We first initialize $t'=\langle \rangle$ and $\Psi=\{a_0\}$. Then we test if $\langle s_0,\Psi,a_1,s_2 \rangle$ is equivalent to  $\langle s_0,a_1,\Psi,s_2 \rangle$. If \emph{true}, $\Psi$ will be $\{a_0,a_1\}$ and $t$ will be $\langle s_0,\Psi,s_2,a_2,...,g \rangle$. If \emph{false}, $t'$ will be $\langle s_0,\Psi \rangle$, $\Psi$ will be $\{a_1\}$ and $t$ will be $\langle s_1,\Psi,s_2,a_2,...,g \rangle$. }
	\item For each plan trace, we generate states that are both partial and noisy by randomly removing a percentage $\xi$ of propositions from complete states and randomly replacing a percentage $\xi$ of remaining propositions in states with other randomly selected propositions. 
	\ignore{
 the probability of exchanging the
order of $a_x$ and $a_y$ is $\frac{\theta_3}{K}$, where $K$ is the 
\emph{distance} between $a_x$ and $a_y$. In the experiments, }
	\ignore{\item For each \emph{proposition} in $\{t'\}$, we ignore it with a $(1-\theta_1)$ probability. If we do not ignore it, we replace it by another different and random \emph{proposition} according the probability of $\theta_2$.}
\end{enumerate}
The real-world domain, called ``CookingTutorial''\footnote{http://cookingtutorials.com/}, which is about how to cook food in the form of natural language. For example, ``\emph{Cook the rice the day before, or use leftover rice in the refrigerator. The important thing to remember is not to heat up the rice, but keep it cold.}'', which addresses the procedure of making egg fired rice. We exploited an off-the-shelf approach {\tt EASDRL} \cite{DBLP:conf/ijcai/FenceZK18} to extract action sequences from the domain. For example, an action sequence of ``\emph{cook(rice), keep(rice, cold)}'' or ``\emph{use(leftover rice), keep(rice, cold)}'' is extracted based on the above-mentioned example. There are 116 texts with 24284 words in total, describing about 2500 actions. Since different action sequences can be extracted from a single text (describing optional ways of cooking in the text), we generated about 400 action sequences, which are generally disordered and noisy (we viewed actions that occur less than 1\% times as noisy actions). In order to learn action models from the action sequences, we built rules based on syntactical parsing of texts to generate initial states and goals for each action sequence. We also manually built action models as ground-truth models to evaluate the model learnt by our approaches. 


\subsection{Criterion}
We define the accuracy $Acc$ of our {\mlp} algorithm by comparing its
learnt action models with the artificial action models which are
viewed as the ground truth. We define the error rate of the learning 
result by calculating the missing and extra predicates of the 
learned action models. Specifically, for each learnt action model 
$a$, if a precondition of $a$ does not exist in the ground-truth 
action model, the number of errors increases by one; if a
precondition of the ground-truth action model does not exist in
$a$'s precondition list, the number of errors also increases
by one. As a result, we have the total number of errors of preconditions
with respect to $a$. We define the error rate of the total number
of errors among all the possible preconditions of $a$, that is, 
\[
Err_{pre}(a)=\frac{\textrm{the\ total\ number\ of\ errors\ of\ preconditions}}
	{\textrm{all\ the\ possible\ precondition\ of}\ a}.
\]
Likewise, we can calculate the error rates of adding effects and
deleting effects of $a$, and denote them as $Err_{add}(a)$ and
$Err_{del}(a)$, respectively. Furthermore, we define the error 
rate of all the action models $\calM$ (denoted as $Err(\calM)$) 
as an average of $Err_{pre}(a), Err_{add}(a)$ and $Err_{del}(a)$ 
for all the actions $a$ in $\calM$, that is,
\[
Err(\calM)=\frac{1}{|\calM |}\sum_{a\in \calM}^{}
\frac{Err_{pre}(a)+Err_{add}(a)+Err_{del}(a)}{3}, 
\]
and define the accuracy as $Acc=1-Err(\calM)$. Note that domain
model $\calM$ is composed of a set of action models.

We compared our {\mlp} to two baselines, {\aman} \cite{DBLP:conf/ijcai/ZhuoK13} and {\arms} \cite{DBLP:journals/ai/YangWJ07}. {\aman} aims to learn action models based on graphical model assuming actions can be noisy in plan traces. {\arms} aims to learn action models based on MAX-SAT framework assuming actions (as well as states) are correct in plan traces.
We evaluated {\mlp} in the following aspects. We first varied the number of plan traces and the rate of observations to see the change of accuracies of our {\mlp} approach, {\aman} and {\arms}, respectively. We then varied the probabilities of disorder and noise to see the change of the three approaches. After that, we evaluated the performance of our approach on the real-world domain with respect to different number of plan traces. Finally, we show the running time to see the efficiency of our {\mlp} approach. 

\ignore{In all experiments we set
$\theta_s=0.1$, $\theta_{diso}=\theta_3$, $w_{do}=w_{po}=w_{no}=5$ and $w_{max}=9999$.}

\subsection{Varying number of plan traces}
\begin{figure*}[!ht]
  \centering
  \includegraphics[width=0.81\textwidth]{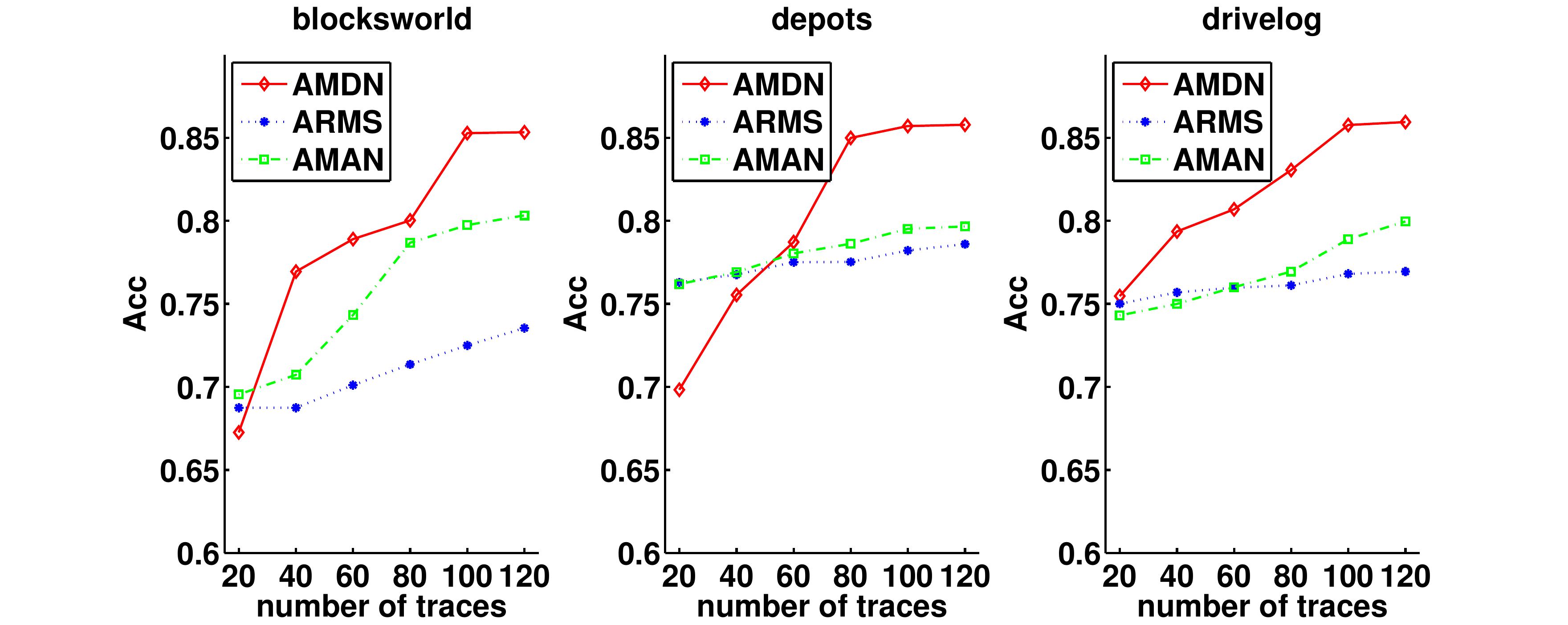}
  \caption{Comparison with respect to number of plan traces}\label{figure:comp}
\mvp
\end{figure*}

\ignore{To see the benefit of our algorithm when solving the disorder and
noisy plan traces, first we generate from 20 to 120 plan
traces with $\theta_1=0.2$, $\theta_2=0.05$ and $\theta_3=0.05$. And} 
We compare the accuracies of action models learnt by {\mlp}, {\aman} and 
{\arms} by varying the number of plan traces. We ran our approach 20 times to calculate an average of accuracies. The results are shown in Figure \ref{figure:comp}. 
\begin{figure*}[!ht]
	\centering
	\includegraphics[width=0.81\textwidth]{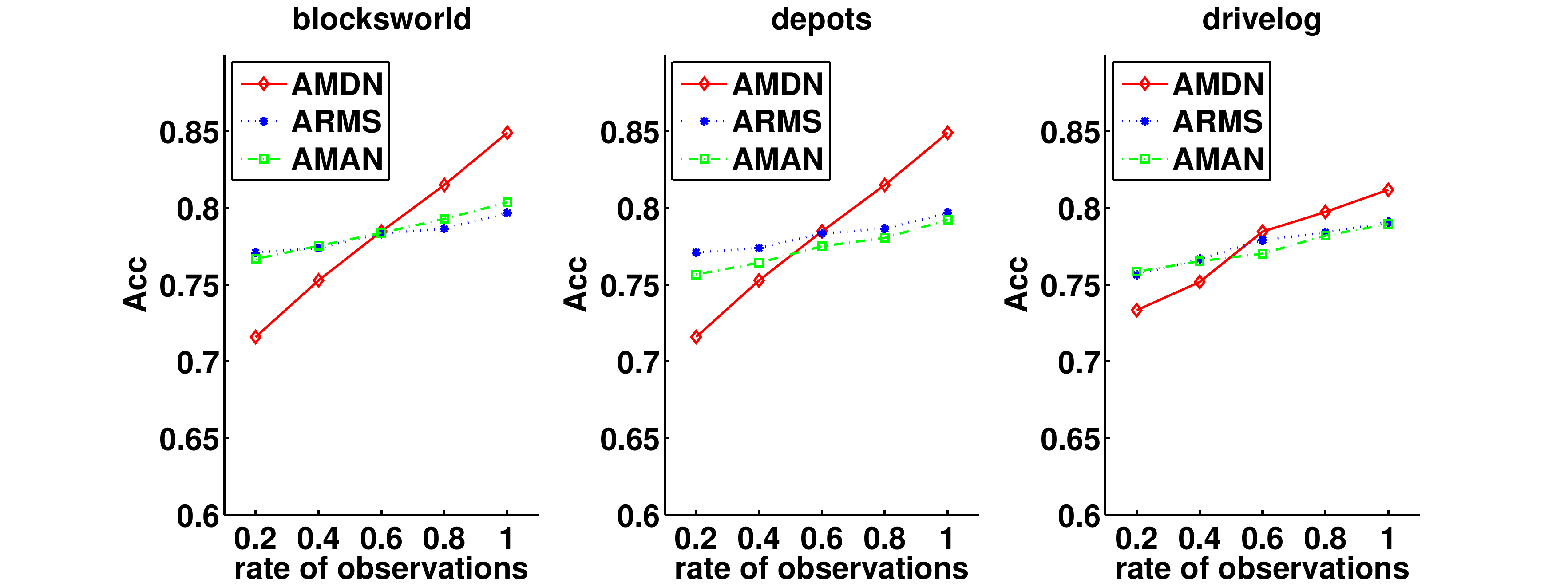}
	\caption{Comparison with respect to rate of observations}\label{figure:comp1}
	\mvp
\end{figure*}

Figure \ref{figure:comp} shows the accuracies of the three
approaches with respect to different number of plan traces. We can see that in all three domains,
the accuracies of {\mlp} are generally
higher than {\aman} and {\arms}, which suggests that {\mlp} is better
at handling disordered and noisy plan traces. This is because
{\mlp} builds more constraints about disorder, noise, \emph{parallel actions}. This is as expected for {\mlp}
has more accurate constraints than {\aman} and {\arms}. On the other hand, the accuracies of {\aman} are generally higher than {\arms} in all three domains. This is because that {\aman} can handle disordered actions by treating them as noise. However, it can only improve the accuracy a little, which suggests that noise has more effect on accuracy than disordered actions. This will be confirmed in the next two experiments (Figures \ref{figure:stod} and \ref{figure:ston}). 
We can also find that as expected, when
the number of plan traces increases, the accuracies are also getting
higher. This is consistent with our intuition, since the more  plan
traces we have, the more information is available for learning domain models of high-quality, and thus helpful for building better action models. 

\subsection{Varying rate of observations}
We compare 
the accuracies of action models learnt by {\mlp}, {\aman} and {\arms} with respect to the rates of observations from 0.2 to 1. We ran our approach 20 times to calcluate an average of accuracy. The results are shown in Figure \ref{figure:comp1}.

From figure \ref{figure:comp1}, we can see that in all three domains, the accuracies of {\mlp}, {\aman} and {\arms}
become higher when the rate of observations increasing, 
which is consistent with our intuition since the more
observations we attain, the more information can be exploited to
improve the learning result. Moreover, when the samples are relatively small, the sample variance may be large, leading {\mlp} to building constraints of bad quality. So the accuracy of {\mlp} is lower than {\aman} and {\arms} when the rate of observations is low. With the increasing of 
the rate of observations, the accuracies of {\mlp} gradually 
become higher than {\aman} and {\arms}. This once again shows that {\mlp}
has higher \emph{learning effectiveness} than {\aman} and {\arms}, it can exploit
more accurate information from the disordered and noisy plan traces.  
 
\subsubsection{Sensitivity to disorder}
\begin{figure}[!ht]
	\hspace{-0.06\textwidth}
	\includegraphics[width=0.59\textwidth]{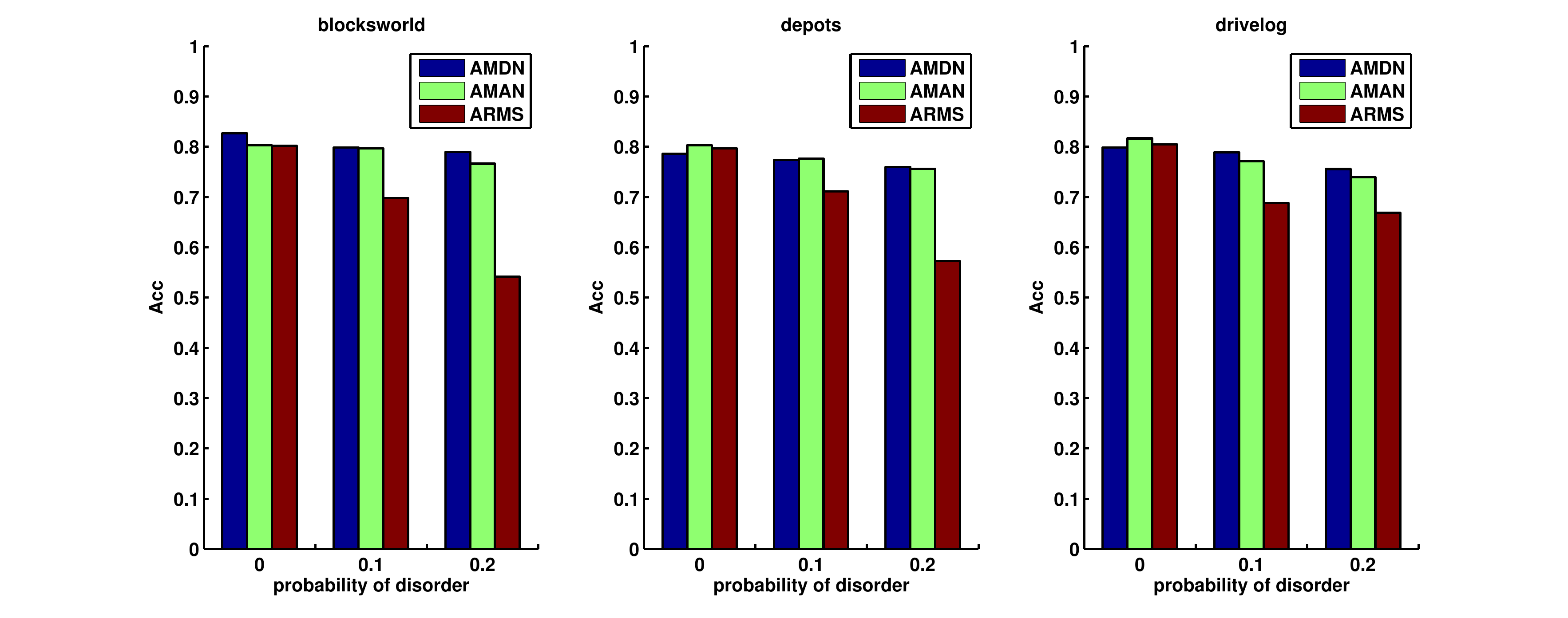}
	\caption{Sensitivity to disorder} \label{figure:stod}
	\mvp
\end{figure}

We also would like to see the sensitivity of {\mlp} with respect to disorder of actions. We ran our approach 20 times to calculate an average of accuracies. The results are shown in Figure  \ref{figure:stod}.

Figure \ref{figure:stod} shows that in all three domains, 
as the probability of disorder increases, the accuracies of all
three algorithms decline, while {\mlp} and {\aman} drop slower than {\arms} 
significantly, which implies {\mlp} and {\aman} are more robust than {\arms} with respect to 
disorder. This is because {\aman} treats disordered actions as noise, and {\arms} does not consider disorder, which results in building many wrong constraints.
{\mlp} builds the constraints about disorder. Although many of
them are wrong, {\mlp} adjusts the weight of each constraint based 
on the probability of disorder, making its constraints more
reasonable. As a result, {\mlp} gets better results than {\arms}.

\subsubsection{Sensitivity to noise}
\begin{figure}[!ht]
  \hspace{-0.06\textwidth}
  \includegraphics[width=0.59\textwidth]{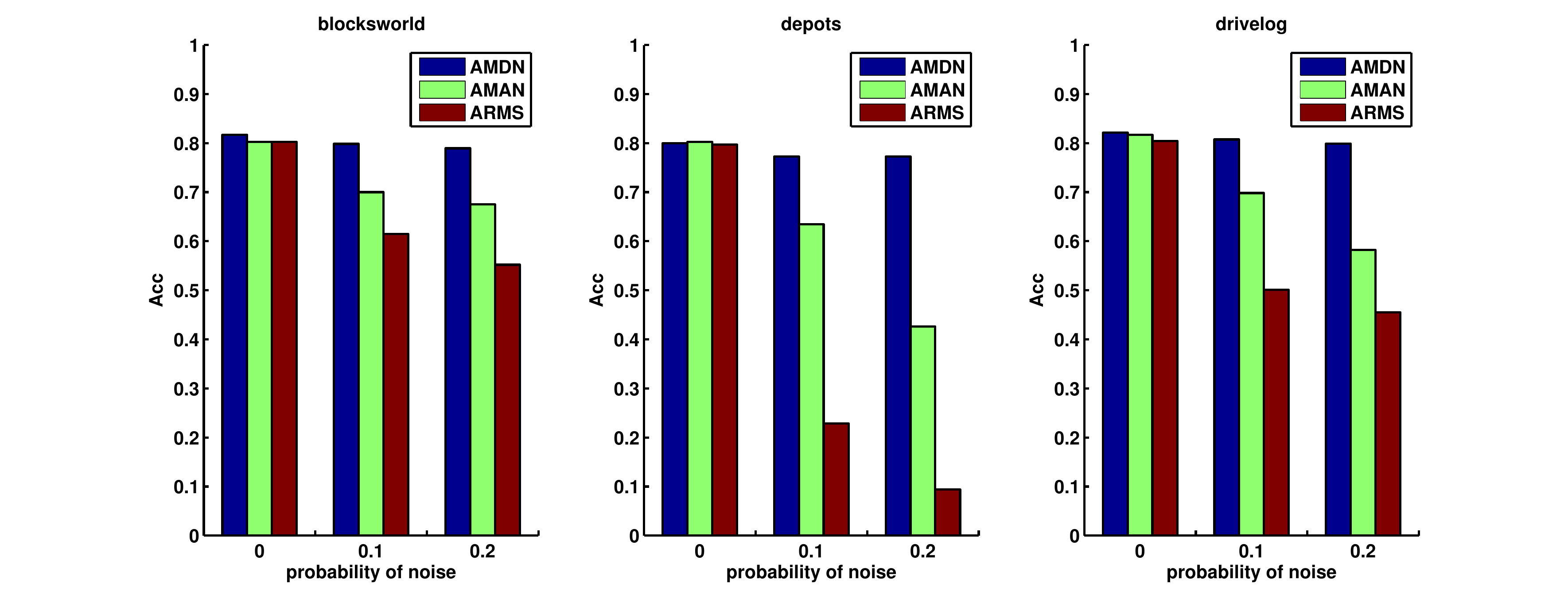}
  \caption{Sensitivity to noise}\label{figure:ston}
\mvp
\end{figure}

We also would like to see the sensitivity of {\mlp} to noise. We ran our approach 20 times to calculate an average of accuracies. The results are shown in Figure \ref{figure:ston}.

The result is shown in Figure \ref{figure:ston}. We find that in
all three domains, as the noise increases, the accuracies of all
three algorithms decline, but {\mlp} drops slower than {\aman} and {\arms} 
significantly, which implies {\mlp} is more robust to 
noise than the other two. Without considering the noise, {\aman} and {\arms} will get a lot of wrong information. In contrast, {\mlp} adjusts the weight of
each constraint based on the probability of noise. Frequently
observed constraints will be given greater weight. So the wrong
constraints will get small weights, which minimizes the effect of
noise on the MAX-SAT solver. 

\subsection{Evaluation on real-world dataset}

\begin{table}[!ht]
\caption{Experimental results on CookingTutorial dataset}\label{NLPdata}
\centering
\begin{tabular}{cccc}
\hline
number of traces & {\tt AMDN} & {\tt ARMS} & {\tt AMAN} \\
\hline\hline
100 & \textbf{0.656} & 0.622 & 0.634  \\
\hline
200 & \textbf{0.788} & 0.651 & 0.682 \\
\hline
300 & \textbf{0.854} & 0.714 & 0.746 \\
\hline
400 & \textbf{0.882} & 0.732 & 0.756 \\
\hline
\end{tabular}
\end{table}

We compared our {\tt AMDN} approach to {\tt ARMS} and {\tt AMAN}. We ran our approach 20 times to calculate an average of accuracies. The experimental results are shown in Table \ref{NLPdata}, where the first column is the number of the plan traces. 

From Table \ref{NLPdata}, we can see that {\tt AMDN} performs much better than both {\tt ARMS} and {\tt AMAN} in all cases, which indicates our constraints built based on noisy and disordered actions can indeed help improve the learning accuracy. We can also find that the accuracy of our approach generally increases with respect to the increase of plan traces. This is consistent with our intuition since the more the plan traces are, the more the knowledge we have for handling disordered and noisy scenarios.

\subsubsection{Running time}
\begin{figure}[!ht]
  \centering
  \includegraphics[width=0.35\textwidth]{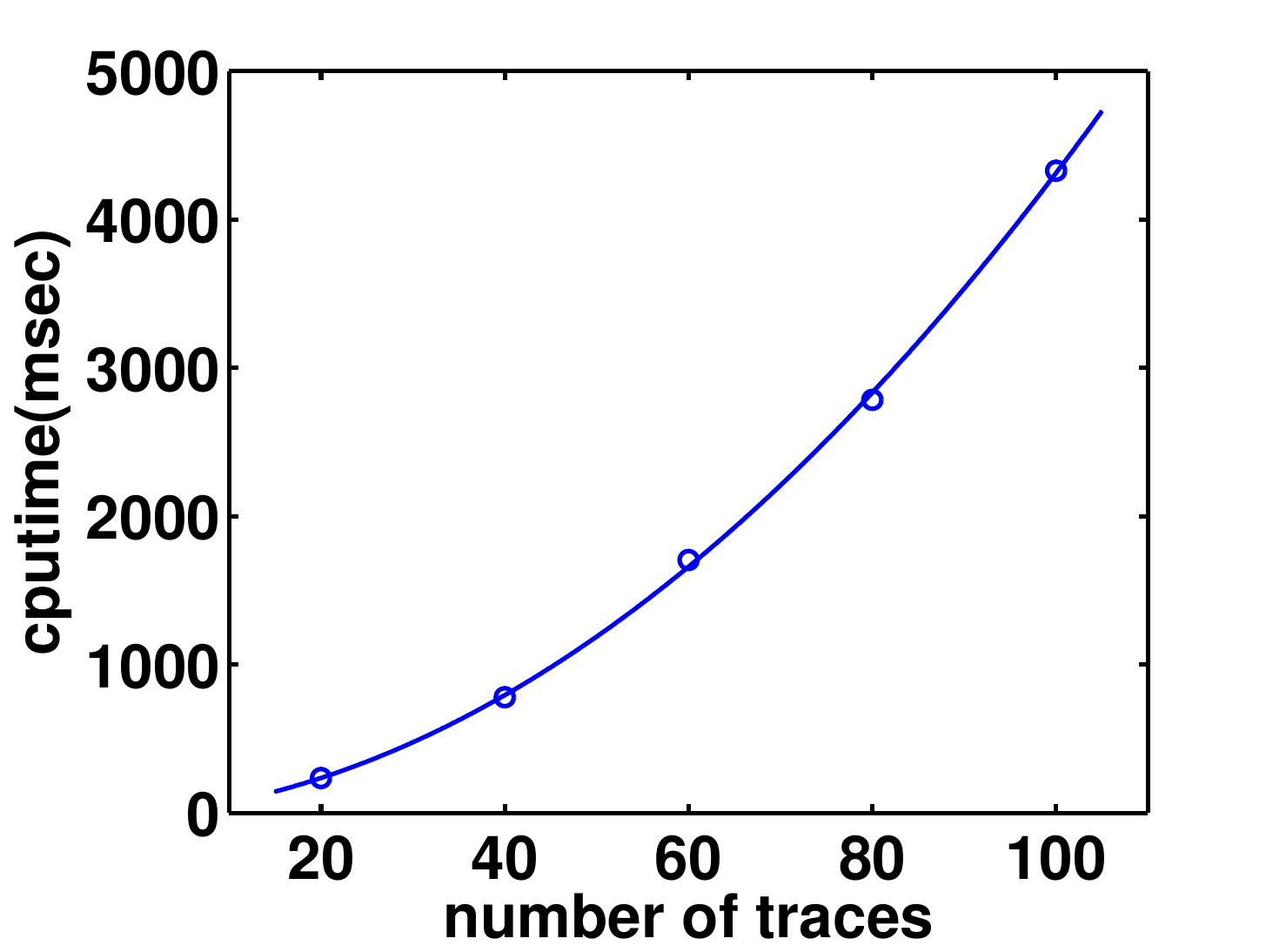}
  \caption{The running time of {\mlp} for domain \emph{blocksworld}.}\label{figure:cputime}
\mvp\mvp
\end{figure}

To study the efficiency of {\mlp}, we ran {\mlp} over 50
problems and calculated an average of the running time with respect to
different number of plan traces in the \emph{blocksworld} domain. The
result is shown in Figure \ref{figure:cputime}. As can be seen from
the figure, the running time increases polynomially with the number of
input plan traces. This can be verified by fitting the relationship
between the number of plan traces and the running time to a
performance curve with a polynomial of order 2 or 3. For example, the
fit polynomial in Figure \ref{figure:cputime} is
$0.3845x^2+4.837x-15.32$. The results for the other two domains are 
similar to \emph{blocksworld}, i.e., the running time also polynomially 
increases as plan traces increase.

\section{Conclusion}
In this paper, we presented a system called {\mlp} for learning domain
models for disordered and noisy plan traces. {\mlp} is able to integrate knowledge
from \emph{parallel actions} and a set of disordered and noisy plan traces to produce
action models. With the plan traces, we first build \emph{disorder constraints}, \emph{parallel constraints} and \emph{noise constraints}, and then
using the weighted MAX-SAT solver to solve them. Our approach is well suited 
for scenarios where high-quality plan traces is hard to attain. Our experiments exhibit that our approach is effective in both planning domains and real-world domain. \ignore{
Different from previous learning systems, our approach allows disordered
\emph{parallel action} and partially observable state with noise in
the plan traces. }

Currently, we consider the disorder of actions in plan traces in between two adjacent actions. It is possible that disorders of actions could happen between distant actions. In the future, it would be interesting to explore the effectiveness of considering distant disorders of actions in plan traces. In addition, we do not consider exploiting training data to learn parameters of probability of disordered actions in Equation (3). In the future, it would be interesting to study how to learn the parameters from training data. Finally,
it would be also interesting to extend {\mlp} 
to investigating the executability of the learnt action models from real-world domains.

\ignore{
\medskip
\noindent {\bf Acknowledgements:} Hankz Hankui Zhuo thanks Natural
Science Foundation of Guangdong Province of China
(No. S2011040001869), Research Fund for the Doctoral Program of Higher
Education of China (No. 20110171120054) and National Natural Science Foundation of China (61033010) for the support of this research. Kambhampati's research is supported  in part by the ARO grant
W911NF-13-1-0023, the ONR grants N00014-13-1-0176, N00014-09-1-0017
and N00014-07-1-1049, and the NSF grant IIS201330813.
}

\newpage
\bibliographystyle{named}
\bibliography{aaai20.bib}

\end{document}